%% file: main.tex
\pdfoutput=1

\documentclass[11pt]{article}

\usepackage{ACL2023}
\usepackage{times}
\usepackage{latexsym}
\usepackage{hyperref}

\usepackage[T1]{fontenc}

\usepackage[utf8]{inputenc}

\usepackage{microtype}

\usepackage{inconsolata}

\usepackage{xspace}
\newcommand{\model}{\textsc{SeqGraph}\xspace}
\newcommand{\fid}{\textsc{FiD}\xspace}
\newcommand{\pathfid}{\textsc{Path-FiD}\xspace}
\newcommand{\hotpot}{\textsc{Hotpot-QA}\xspace}
\newcommand{\musique}{\textsc{Musique}\xspace}
\newcommand{\roberta}{\textsc{RoBERTa}\xspace}
\newcommand{\longformer}{\textsc{Longformer}\xspace}
\usepackage{float} 
\usepackage{graphicx}
\usepackage{amsmath}
\usepackage{amsfonts}
\usepackage{microtype}
\usepackage{booktabs} 
\usepackage{xspace}
\usepackage{subcaption}
\usepackage{bbm}
\usepackage{multirow}
\usepackage{enumitem}
\usepackage{xspace}
\usepackage{cleveref}
\usepackage{adjustbox}
\usepackage{kky}
\usepackage{adjustbox}
 
\usepackage[textsize=scriptsize]{todonotes}
	
\usepackage{color, colortbl}
\usepackage{arydshln}

\newcommand{\Scref}[1]{\S\ref{#1}}

\newcommand{\ie}{\textit{i}.\textit{e}.}
\newcommand{\eg}{\textit{e}.\textit{g}.}
\newcommand{\ph}[1]{[\texttt{#1}]}

\definecolor{applegreen}{rgb}{0.55, 0.71, 0.0}
\definecolor{amaranth}{rgb}{0.9, 0.17, 0.31}
\definecolor{copperrose}{rgb}{0.6, 0.4, 0.4}
\definecolor{alizarin}{rgb}{0.82, 0.1, 0.26}

\newcommand{\bluett}[1]{\textcolor{blue}{#1}}
\newcommand{\orangett}[1]{\textcolor{orange}{#1}}
\newcommand{\greentt}[1]{\textcolor{applegreen}{#1}}
\newcommand{\browntt}[1]{\textcolor{copperrose}{#1}}
\newcommand{\redtt}[1]{\textcolor{alizarin}{#1}}
\title{Single Sequence Prediction over Reasoning Graphs for Multi-hop QA}

  \author{Gowtham Ramesh\thanks{Equal contribution},  Makesh Sreedhar\footnotemark[1],   and Junjie Hu \\
  University of Wisconsin-Madison \\
  \texttt{\{gramesh4,msreedhar,junjie.hu\}}@wisc.edu 
  }

\begin{document}
\maketitle
\begin{abstract}

Recent generative approaches for multi-hop question answering (QA) utilize the fusion-in-decoder method~\cite{izacard-grave-2021-leveraging} to generate a single sequence output which includes both a final answer and a reasoning path taken to arrive at that answer, such as passage titles and key facts from those passages. While such models can lead to better interpretability and high quantitative scores, they often have difficulty accurately identifying the passages corresponding to key entities in the context, resulting in incorrect passage hops and a lack of faithfulness in the reasoning path. To address this, we propose a single-sequence prediction method over a local reasoning graph (\model)\footnote{Code/Models will be released at \url{https://github.com/gowtham1997/SeqGraph}} that integrates a graph structure connecting key entities in each context passage to relevant subsequent passages for each question. We use a graph neural network to encode this graph structure and fuse the resulting representations into the entity representations of the model. Our experiments show significant improvements in answer exact-match/F1 scores and faithfulness of grounding in the reasoning path on the HotpotQA dataset and achieve state-of-the-art numbers on the Musique dataset with only up to a 4\% increase in model parameters.

\end{abstract}

\input{sections/01-introduction.tex}

\input{sections/02-methods.tex}

\input{sections/03-experiments.tex}

\input{sections/04-results.tex}

\input{sections/05-analysis.tex}

\input{sections/06-related.tex}

\input{sections/07-conclusion.tex}

\section*{Limitations}



We identify the following limitations of our work:

\paragraph{Longer Output Sequences} While outputting the reasoning path as a single short sequence makes the model more interpretable, it increases the challenge of producing a long /coherent sequence when the question is complex (more than 3 hops). Producing a longer sequence also increases the inference time. Simplifying this output while  not sacrificing interpretability is a good future direction

\paragraph{Entity Identification}
Our method needs wikipedia outlinks or a entity linker to construct a localized graph for every question. Generalizing this step by pretraining the model to do entity linking \cite{Fvry2020EntitiesAE, pmlr-v139-sun21e, Verga2020FactsAE} might eliminate the need to use an external module.

\bibliography{anthology,custom}
\bibliographystyle{acl_natbib}
\clearpage
\newpage
\onecolumn
\appendix

\input{sections/appendix}

\end{document}

%% file: sections/01-introduction.tex
\section{Introduction}

Multi-hop Question Answering (QA) involves reasoning over multiple passages and understanding the relationships between those pieces of information to answer a question. Compared with single-hop QA, which often extracts answers from a single passage, multi-hop QA is more challenging as it requires a model to determine the relevant facts from multiple passages and connect those facts for reasoning to infer the final answer.

\begin{figure}[t]
    \centering
    \includegraphics[width=\columnwidth, height=6cm]{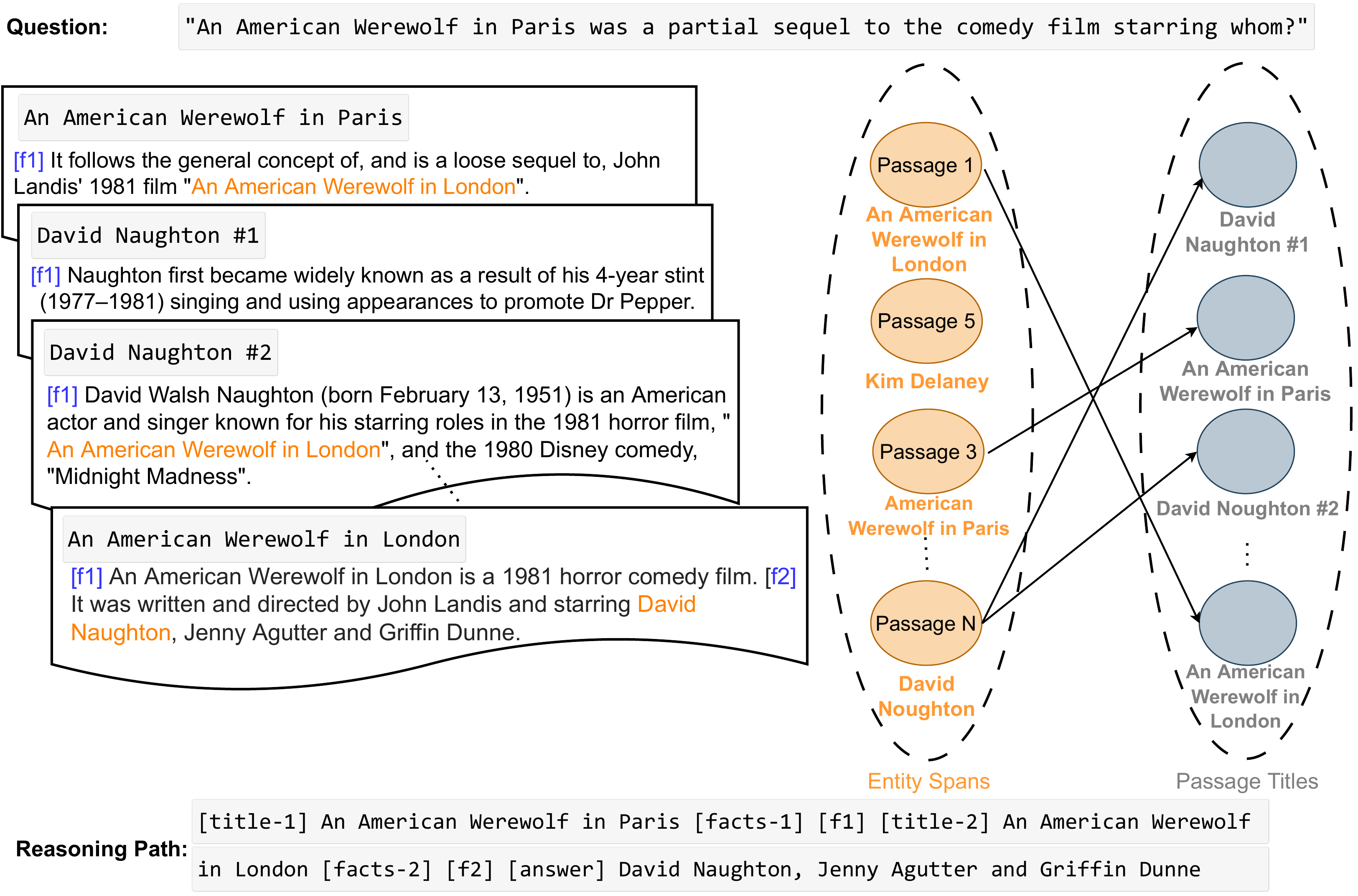}
    \caption{Localized graph construction connecting entity spans to corresponding passages in the context. If there are multiple passages with the same title, we connect the entity span to all such passages. }
    \label{fig:graph_construction}
    \vspace{-8mm}
\end{figure}

To tackle multi-hop QA, recent works have investigated large pretrained \textit{generative} models~\cite{DBLP:conf/nips/LewisPPPKGKLYR020, roberts-etal-2020-much, NEURIPS2020_1457c0d6} and demonstrated their effectiveness over traditional \textit{extractive} models~\cite{chen-etal-2017-reading}. Compared with extractive models, the ability of generative models to effectively aggregate and combine evidence from multiple passages proves advantageous for multi-hop QA. In particular, \citet{izacard-grave-2021-leveraging} propose a method called \fid (Fusion-in-Decoder), which leverages passage retrieval with a generative model, such as T5~\cite{raffel2020exploring} or \textsc{BART}~\cite{lewis-etal-2020-bart}, to achieve state-of-the-art performance on various single-hop QA tasks. However, this approach does not extend well to multi-hop QA tasks~\cite{yavuz-etal-2022-modeling}, as it sorely relies on a black-box generative model to generate answers directly without explicitly modeling the multi-hop reasoning process. Additionally, \fid encodes multiple context passages independently for multi-hop QA, ignoring the structural and semantic relationship between these passages~\cite{yu-etal-2022-kg}. Building on \fid, \pathfid~\cite{yavuz-etal-2022-modeling} addresses the interpretability issue by training a model to generate a reasoning path that contains supporting passage titles, facts, and the final answer. However, our analysis of \pathfid outputs shows \textit{disconnected reasoning} with incorrect passage hops in the model's reasoning path which affects final answer generation. Recently, there have been multiple techniques~\cite{jiang-bansal-2019-avoiding, lee-etal-2021-robustifying, ye-etal-2021-connecting} to counter disconnected reasoning which operate at the dataset level, using adversarial training, adding extra annotations or using dataset rebalancing for training. While these approaches optimize models to mitigate disconnected reasoning~\cite{trivedi-etal-2020-multihop}, the performance on the original test set often suffers from a significant decrease. 

In this paper, we propose a single-\textbf{seq}uence prediction method over a local reasoning \textbf{graph} (\model) that integrates a graph structure connecting key entities in each context passage to relevant subsequent passages for each question. Different from the prior works, our method not only mitigates the disconnected reasoning issue but also maintains robust performance on the original dataset. Intuitively, for each multi-hop question, our method leverages the structural relationship between different passages to learn structured representations through a graph neural network (GNN)~\cite{NIPS2017_5dd9db5e, kipf2017semi}. The structured representations are fused to bias the generative model toward predicting a faithful, connected reasoning path which improves answer predictions. Our experiments on the \hotpot dataset~\cite{yang2018hotpotqa} show clear improvements in exact-match(EM)/F1 scores compared to generative baselines in the \textit{distractor} setting while minimizing disconnected reasoning quantified by the \textsc{DiRe} score~\cite{trivedi-etal-2020-multihop}. We also achieve the state-of-the-art performance on the \musique-Answerable test dataset~\cite{10.1162/tacl_a_00475} with a 17-point improvement in answer F1 over the current best-performing model in the end-to-end (E2E) category.  

To summarize, our contributions are as follows:

\begin{itemize}[leftmargin=10pt]\itemsep-0.2em
    \item We propose an interpretable single-\textbf{seq}uence prediction approach over local reasoning \textbf{graph}s, \model, to bias the model representations
    \item \model achieves notable performance improvements on two multi-hop QA benchmarks, \hotpot and \musique (SOTA), with only a minimal increase in the model size.
    \item \model reduces disconnected reasoning as measured by \textsc{DiRe} score while maintaining strong performance gains on the original dataset.
\end{itemize}

%% file: sections/02-methods.tex
\section{Preliminaries}
\label{sec:preliminaries}

\paragraph{Problem Setup:} In a multi-hop QA task, each QA pair in a labeled dataset $\Dcal$ is given along with a set of $N$ passages, $\Pcal_q = \{p_1, p_2, . . . , p_N\}$, \ie, $(q, a, \Pcal_q)\in \Dcal$, where a passage has its title and content $p_i=(t_i, c_i)$. The task is to learn a model parameterized $\theta$ to generate an answer string $a$ for the given question $q$ and $\Pcal_q$. 

In this paper, we focus on the \textit{distractor} setting, where $\Pcal_q$ is given for each question and contains $m$ distractors that are not useful to the answer prediction. Thus, this task requires a model to reason over multiple hops of the remaining ${N-m}$ relevant passages. In addition to predicting the final answer $a$, we also aim to train a model to predict a \textit{reasoning path} $R$ of important elements (\eg, relevant passage titles, supporting facts in a passage) that lead to the final answer.  

\paragraph{Multi-hop QA as Single Sequence Generation:} Recent generative question answering (QA) approaches (e.g., \fid~\cite{izacard-grave-2021-leveraging}, \pathfid~\cite{yavuz-etal-2022-modeling}) utilize an encoder-decoder model as the backbone to generate answers in a single text sequence. In particular, \fid is one of the popular formulations. 

Specifically, for each passage $p_i=(t_i, c_i)\in\Pcal_q$ of a question $q$, \fid encodes a combined sequence of the question, the passage title and contents into an embedding. These embeddings for all passages are concatenated as inputs to the decoder for generating the final answer.

\pathfid builds upon this by explicitly modeling a reasoning path as part of the generation output in addition to the answer. Specifically, special index tokens $[f_i]$ are added to demarcate all sentences in each passage context. The sentences supporting the prediction of a final answer are considered facts. The decoder is then trained to generate the reasoning path $R$ as a linearized sequence consisting of the passage titles and the index tokens of facts used within those passages to obtain the final answer. Figure~\ref{fig:graph_construction} shows an example of a reasoning path.

\paragraph{Disconnected Reasoning in \pathfid:} Since the model predictions now include the reasoning path, we can analyze which facts in the passage are utilized by the model to determine the next passage to hop to and arrive at the final answer. 
For a perfectly faithful model,  all predictions with correct answers should have correctly identified passages and facts. However, due to the presence of shortcuts in the datasets as well as the model's predicted reasoning path not being faithful, 
we observe model predictions containing correct final answers but incorrect identification of passage titles or facts. This unfaithful prediction issue is referred to as \textit{disconnected reasoning}~\cite{trivedi-etal-2020-multihop}. Different from \pathfid, we use the presence of a local graph structure between different passages in the context to bias the representations of the model and help alleviate this problem.

\section{Method}

In this section, we describe our proposed method for solving disconnected reasoning for multi-hop QA in the \textit{distractor} setting.

\paragraph{Overview:} Our method first constructs a local graph over passage contexts for each question (\Scref{sec:graph-construction}), and integrates the graph information with the key entities to improve the generation of reasoning paths (\Scref{sec:entity_passage_fusion}). Different from prior works that encode all the passages independently, we connect the passages through the key pivot entities into a local graph for a question, which allows us to encode structural representations across passages by a graph neural network. These graph structured representations are then fused with the contextualized text representations from a text encoder, guiding the model to leverage structural information to alleviate disconnected reasoning over passages.

\subsection{Graph Construction}
\label{sec:graph-construction}
In contrast to the \textit{full-wiki} setting where a model must retrieve relevant passages from Wikipedia or a large corpus, the distractor setting provides the model with a list of $N$ passages $\Pcal_{q}$ consisting of $N-m$ relevant passages and $m$ distractors for each question $q$. Conventionally, these passages are collected from Wikipedia, as Wikipedia remains one of the largest faithful knowledge sources available for public usage. Even for text passages out of Wikipedia, there are existing out-of-box entity linkers (e.g., SLING~\cite{ringgaard2017sling}, BLINK~\cite{wu-etal-2020-scalable}) that can identify key entities from texts and link them to their Wikipedia pages. As a result, each provided passage may contain pivot entities with hyperlinks connecting to their corresponding Wikipedia pages. We exploit such entity hyperlinks to construct a local directed graph $\Gcal=(\Ncal, \Lcal)$ containing two types of nodes (\ie, entities and passage titles) and links between these nodes. Specifically, for each pivot entity $e$ in a passage $p_i$, we create a link from $e$ to the title $t_j$ of another passage $p_j$ (denoted as $l_{e\rightarrow t_j}$) whenever the entity span $e$ points to a Wikipedia article that contains the passage $p_j$. 

For example, an entity span \textit{``David Noughton''} appears in the passage context: \textit{``An American Werewolf in London is a 1981 horror comedy film starring David Noughton, Jenny Agutter. ...''}

This entity would be connected to a passage with the title of \textit{``David Walsh Noughton''}, forming the link (David Noughton[Entity] $\rightarrow$ David Walsh Noughton[Passage]). If there are multiple passages with the title \textit{``David Walsh Noughton''} among the $N$ passages, the entity span would be connected to all of them with distinct links. Figure \ref{fig:graph_construction} shows an example of an entity-passage graph.

\subsection{Entity-to-Passage Fusion}
\label{sec:entity_passage_fusion}

Next, we describe how we encode such a local directed graph into vector representations for all nodes and fuse these node representations with the contextualized text representations of the corresponding entities from the language model.  
\begin{figure*}[t]
    \centering
    \includegraphics[width=\textwidth, height=11cm, trim={0 3cm 0 0 }, clip]{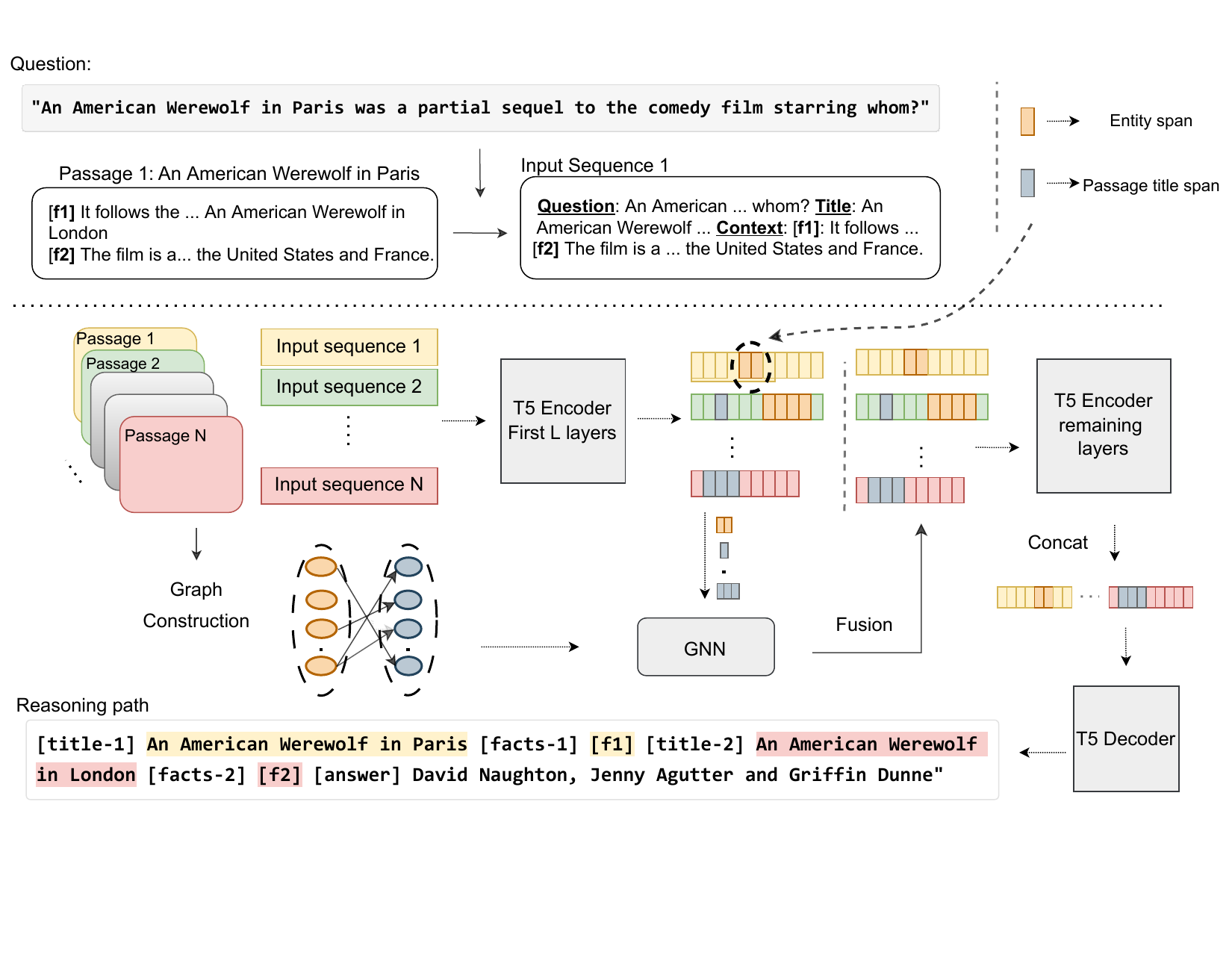}
    \caption{Given a question and the supporting passages, we construct a localized entity-passage graph. The representations from the $L^{th}$ layer of the language model is used to initialize the weights of a graph neural network(GNN) and it is used to perform message passing on the constructed local graph. The representations for the entity spans and titles from the GNN are added to the LM representations and passed through the remaining $M-L$ layers of the encoder. The T5 decoder performs cross-attention on the final hidden states from the encoder and generates the reasoning path with the final answer.}
    \label{fig:seqgraph_main}
\end{figure*}

We utilize the same model as \pathfid with a pre-trained T5 model as our backbone architecture. The input for this method consists of the $N$ sequences, where each sequence is a concatenation of the question $q$, the title and contents of a passage $p_i$ from the collection $p_i\in\Pcal_q$ together with their indicator tokens, denoted as $S_i$ below: 
\begin{align}
S_i := \ph{Question}~q~\ph{Title}~t_i~\ph{Content}~c_i
\end{align}

Given the T5's encoder of $M$ transformer layers, we first encode $S_i$ through the first $L$ layers to obtain the intermediate hidden representations $\Zb_i^L$ in Eq.~(\ref{eq:text_encoder_L}), which capture the shallow contextualized information of the input sequence. 
\begin{align}\label{eq:text_encoder_L}
\Zb_i^{L} = \text{TextEncoder}(S_i, L)
\end{align}

We utilize these shallow representations to initialize the node embeddings for a graph neural network. Specifically, we extract the representations of the entity spans or passage title spans (\ie, nodes in the graph $\Gcal$) from $\Zb_i^L$ according to their span positions $[a,b]$ in $S_i$. Next, for a text span $S_{i, a:b}$ representing either an entity or a title in $S_i$, we average the extracted representations of the text span to obtain an initial node embedding, \ie, $\nbb = \text{avg}(\Zb_{i, a:b}^L)$. Finally, we stack the initial embeddings for all nodes denoted as $\Nb$ and apply a graph neural network (GNN) to further encode the structural embeddings on the graph $\Gcal$:
\begin{align}
    \Zb^G = \text{GraphEncoder}(\Nb, \Gcal)
\end{align}

As we record the text span position $[a, b]$ for each node in $\Gcal$, we can leverage the node embeddings $\Zb^G$ to construct a new structured representation $\Zb_i^G$ (with the same size as $\Zb_i^L$) for each sequence $S_i$ where we fill in the node embeddings from $\Zb^G$ to their corresponding text span positions $[a, b]$ in $S_i$ and fill in $0$ to the other non-span positions. 

Finally, we fuse the contextualized text representations $\Zb_i^L$ from the text encoder and the structured node representations $\Zb_i^G$ by an aggregation operator $\oplus$, and pass them to the remaining layers of the text encoder to obtained the fused representations $\Sbb_i$ for each input sequence $S_i$: 
\begin{align}
\Sbb_i = \text{TextEncoder}(\Zb_i^G \oplus \Zb_i^{L}, M-L)
\end{align}
In this work, the aggregation operator used is a simple addition. Complex aggregation mechanisms such as learning a weighted combination of the representations can be explored in future work.

We concatenate the fused representations $\Sbb_i$ from all of the $N$ context sequences to form $\Sbb=\left[\Sbb_{1}; \Sbb_2 \cdots ; \Sbb_{N}\right]$.  

 Subsequently, $\Sbb$ is passed as inputs to the T5 decoder that estimates the conditional probability $P_\theta(R|\Sbb)$ of predicting a reasoning path $R$. Depending on the annotations in different datasets, a reasoning path $R$ can take various formats. For example, the reasoning path takes the form ``$R:= \ph{title}~t_i ~\ph{facts}~f_i ~\ph{answer}~a$''
 for \hotpot and ``$R:= \ph{title}~t_i ~\ph{intermediate\_answer}$ $\text{ans}_i~\ph{answer}~a$'' for \musique. We also investigate variants of reasoning paths for \musique in our experiments.
 As we can construct ground-truth reasoning paths $R^*$ during training, the model is optimized using a cross-entropy loss between the conditional probability $P_\theta(R|\Sbb)$ and $R^*$.

%% file: sections/03-experiments.tex
\section{Experimental Setting}
\label{sec:exp:dataset}
In this section, we elaborate on the datasets, the baseline models and the variants of \model we consider for our experiment settings.
We consider two multi-hop QA datasets, \hotpot and \musique. Since \model is primarily focused only on improving the efficacy of encoding, we consider only the \textit{distractor} setting for both datasets. Table \ref{tab:dataset-stats} shows the standard train/dev/test statistics.

\paragraph{\hotpot:}

The final answer to each question in the distractor setting is extracted from 10 passages. The dataset includes two main types of questions: bridge (80\%) and comparison (20\%). Bridge questions often require identifying a bridge entity in the first passage to correctly hop to the second passage that contains the answer, while comparison questions do not have this requirement. Each question is also provided with annotations of 2 supporting passages (2-hop) and up to 5 corresponding relevant sentences as their supporting facts.

\paragraph{\musique:} \musique has questions that range in difficulty from 2 to 4-hops and six types of reasoning chains. \musique uses a stringent filtering process as well as a bottom-up technique to iteratively combine single-hop questions from several datasets into a $k$-hop benchmark that is more difficult than each individual dataset and significantly less susceptible to the disconnected-reasoning problem. Unlike \hotpot, \musique does not provide annotations of relevant sentences but provides supporting passage titles, question decomposition(decomposition of a multi-hop question into simpler 1-hop sub-questions) and also intermediate answers to the decomposed questions. Given this variety, we use the following reasoning path variants to train the model to generate:

\begin{itemize}[leftmargin=15pt]\itemsep-0.2em
    \item DA: Question decomposition and final answer
    \item SA: Supporting titles and final answer
    \item SIA: Supporting titles, intermediate answers and final answer
    \item DSIA: Question decomposition, supporting titles, intermediate answers and final answer
\end{itemize}

Table \ref{tab:Reasoning-path-variants} shows an example of different reasoning paths. While the last variant (predicting every decomposition/intermediate answer or support title) is more interpretable, it encounters the challenge of producing a long sequence. SIA is our best-performing reasoning path variant which is used for all of our results and analysis.

\subsection{Models in Comparison}

Our main baselines are generative approaches to multi-hop QA that include and build upon the \fid approach. For all of the models, we use the pre-trained T5 encoder-decoder as the backbone and consider two sizes---base and large variants.

\begin{itemize}[leftmargin=15pt]\itemsep-0.2em
    \item \fid: Model generation includes only the final answer.
    \item \pathfid: Model generation includes the reasoning path as well as the final answer.
    \item \model: Model that utilizes a fusion of representations from the language model and the Graph Neural Network. Similar to \pathfid, we train the model to generate the reasoning path in addition to the final answer.
\end{itemize}

\subsection{Evaluation Metrics}
For both \hotpot and \musique, we use the standard quantitative metrics of exact-match and F1 scores to evaluate the quality of predicted answers. For models that predict the reasoning path in addition to the final answer, we can quantify how accurately they can identify the supporting facts (or supporting titles for \musique) using the Support-EM and Support-F1 scores~\citet{yang2018hotpotqa}. 

To quantify the level of disconnected reasoning, we compute dire F1 scores on the answer spans (\textbf{Answer}), supporting paragraphs (\textbf{Supp}$_\text{p}$), supporting sentences (\textbf{Supp}$_\text{s}$), joint metrics (\textbf{Ans+Supp}$_\text{p}$, \textbf{Ans+Supp}$_\text{s}$) of the Dire \hotpot subset.
\subsection{Implementation details}

We train all models using an effective batch size of 64. We use an initial learning rate of 1e-4, a linear rate scheduler, a warmup of 2,000 steps (1,000 steps for \musique), and finetune the models for 10 epochs. For \model, we use GAT \citep{https://doi.org/10.48550/arxiv.1710.10903} for our GNN layers. A maximum sequence length of 256 tokens is used for constructing the input.  All experiments have been conducted on a machine with either 4$\times$40G A100 GPUs or 4$\times$80G A100 GPUs. A detailed list of hyperparameters can be found in Appendix \ref{sec:detail-hyperparam}.

%% file: sections/04-results.tex
\input{Tables/multi-hop-main.tex}

\section{Results and Analysis}
\label{sec:result}

In this section, we present the main results of the baselines and our proposed approach on \hotpot and \musique (\Scref{sec:result:main}), and then perform fine-grained analysis thereafter.

\subsection{Multi-hop Performance}
\label{sec:result:main}

The quantitative performance of the models in terms of exact-match and F1 scores for both the final answer and the predicted supports are shown in Table \ref{tab:dev-results}. We find that across both model sizes (\textsc{Base} and \textsc{Large}), explicitly predicting the reasoning path helps \pathfid in improving the answer EM and F1 scores over the vanilla \fid approach. By biasing the model with graph representations, \model outperforms the baselines on both the \hotpot and the \musique datasets. 

\model achieves a 2-point improvement in both answer and support EM when considering the base variant and ~1.5 point improvement for the large variant on the dev set of \hotpot.

On the more challenging \musique dataset, we observe stronger results from \model where it records up to a 4-point improvement in both answer and support scores across both model sizes on the dev set. On the test set (in Table \ref{tab:musique-leaderboard} of the appendix), the current best performing approach is a two stage \roberta / \longformer-Large model, Select-Answer, where the passage selection/ranking and answer generation stage is optimized separately using different models. \model-Large achieves state-of-the-art numbers on Answer-F1 with a 5-point improvement over the Select-Answer model\footnote{https://leaderboard.allenai.org/musique\_ans/} even though it is a single stage approach. When comparing with the top score in the end-to-end (E2E) category which all of our models belong to, \model gets a massive 17-point improvement in answer F1 and a 9-point improvement in support F1 establishing the efficacy of our approach. It should also be noted that all of the current models on the leaderboard are discriminative approaches with an encoder-only model (\longformer-Large) encoding a very long context length of 4,096, while all of our models are generative in nature with a much smaller context length of 256. \musique is also designed to be more challenging than \hotpot and explicitly tackles the issue of disconnected reasoning during dataset curation, making it harder for the model to take shortcuts and cheat. The larger performance improvements of \model on \musique compared to \hotpot showcases the advantage of our proposed approach, providing promising results for further research in this direction to mitigate disconnected reasoning.

%% file: Tables/multi-hop-main.tex
\begin{table*}[ht!]
\centering
\small
\begin{tabular}{@{}lcccccccc@{}}
\toprule
\multirow{2}{*}{Model} & \multicolumn{4}{c}{\hotpot} & \multicolumn{4}{c}{\musique}  \\ \cmidrule{2-9}
 & \multicolumn{2}{c}{Answer} & \multicolumn{2}{c}{Support} & \multicolumn{2}{c}{Answer} & \multicolumn{2}{c}{Support} \\ 
 & EM & F1 & EM & F1 & EM & F1 & EM & F1 \\ \midrule
\fid-Base & 61.84 & 75.20 & - & - & 29.38 & 39.97 & - & - \\
\pathfid-Base & 62.03 & 75.69 & 60.45 & 86.00 & 34.71 & 44.93 & 57.30 & 80.18 \\
\model-Base & \textbf{64.19} & \textbf{77.60} & \textbf{62.44} & \textbf{87.72} & \textbf{37.36} & \textbf{47.11} &\textbf{ 58.05} & \textbf{80.39} \\
\midrule
\fid-Large & 65.59 & 79.39 & - & - & 36.04 & 46.66 & - & - \\
\pathfid-Large$^{*}$ & 65.80 & 78.90 & 59.30 & 85.70 & - & - & - & -\\
\pathfid-Large & 65.33 & 79.00 & 61.52 & 86.88 & 42.28 & 53.86 & 62.14 & 82.45 \\
\model-Large  & \textbf{66.51} & \textbf{81.62} &\textbf{ 63.24} & \textbf{88.28} & \textbf{46.01} &\textbf{ 56.88} & \textbf{65.12} & \textbf{83.65} \\ 
\bottomrule

\end{tabular}

\caption{Performance on the dev set of \hotpot and \musique. Since \fid does not predict a reasoning path, we do not compute the Support EM and F1 scores. \pathfid-Large$^{*}$ indicates the numbers reported from \citet{yavuz-etal-2022-modeling}, while the other numbers are from our reimplementation}
\label{tab:dev-results}
\end{table*}

%% file: sections/05-analysis.tex
\label{sec:result:faithfulness}
\begin{figure*}[htbp]
    \centering
    \includegraphics[width=12cm, height=6.5cm]{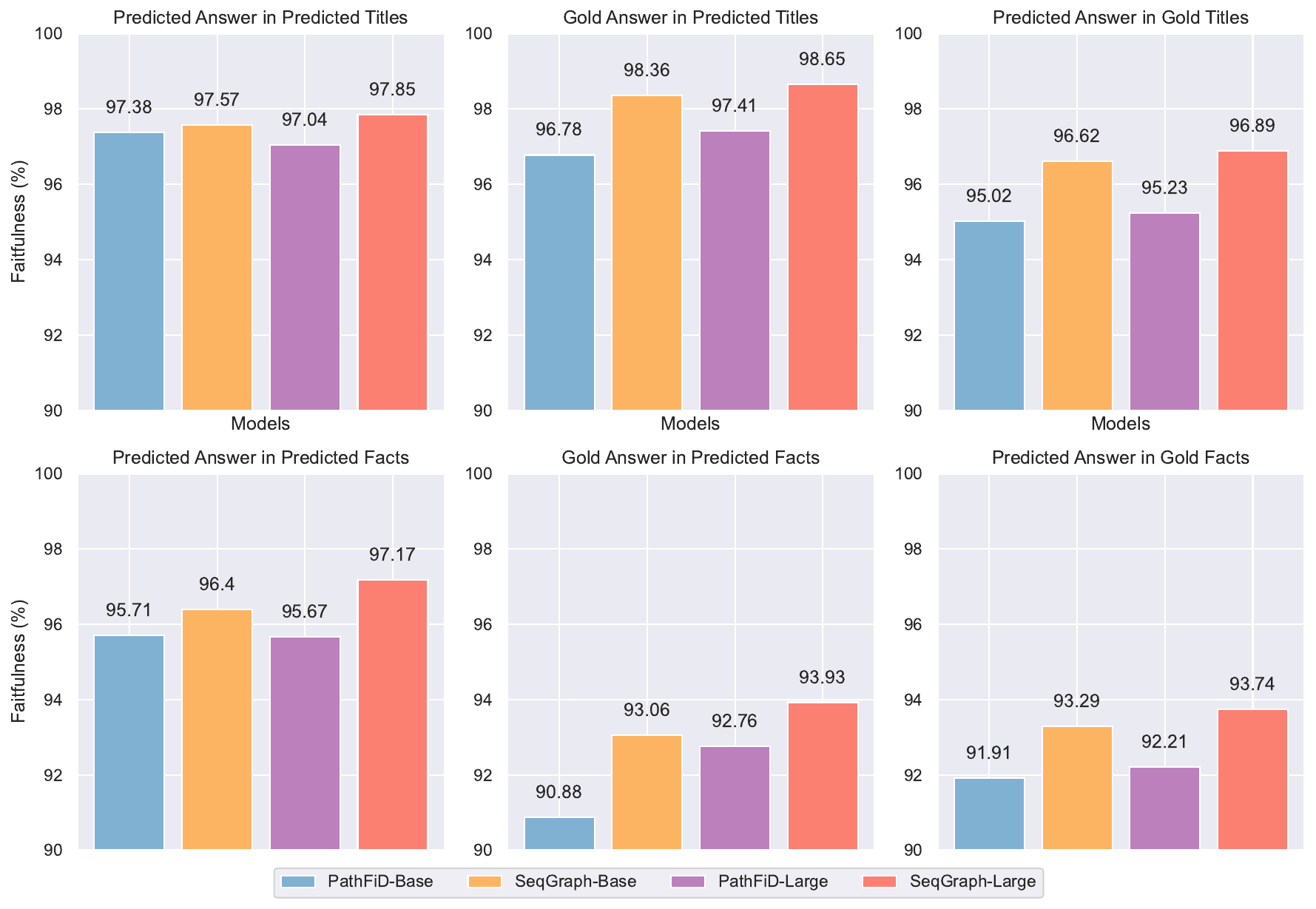}
    \caption{Comparison of model faithufulness on \hotpot. We find that \model improves over \pathfid consistently across all categories.}
    \label{fig:hotpot_faithful}
\end{figure*}
\input{Tables/dire_score.tex}

\subsection{Faithfulness of Reasoning Paths}

We follow \citet{yavuz-etal-2022-modeling} to perform analysis at the passage and individual fact level to determine how faithful the generated reasoning paths are across different models. 

\paragraph{Predicted Answer in Predicted Titles/Support:}\textit{how often are the predicted answers found in one of the predicted passages or in the predicted supporting facts\footnote{We do this analysis only on Bridge type questions where the final answer span can be found in context passages, unlike comparison questions where the final answer is usually \textit{yes/no}}}.

\paragraph{Gold Answer in Predicted Titles/Support:}\textit{how often are the gold answers found in one of the predicted passages or in the predicted supporting facts}.

\paragraph{Predicted Answer in Gold Titles/Support:}\textit{how often are the predicted answers found in one of the gold passages or in the gold supporting facts.}

Figure \ref{fig:hotpot_faithful} shows the described faithfulness metric scores on \hotpot. We find that \model is more faithful with a 0.5-1.5\% improvement over \pathfid across all considered categories.

\subsection{Performance vs Number of hops}

We break down the final answer exact-match and F1 scores based on how many supporting facts(or titles for Musique) are required to answer the question. Figure \ref{fig:hotpot_num_hops} shows this performance breakdown for \hotpot and Figure \ref{fig:musique_num_hops} shows it for \musique. We observe that \model improves over \pathfid in the cases where the support includes two or three supporting facts (or titles), but the answer EM takes a hit when the number of supporting facts(titles) $\geq 4$. We notice that \model has a higher support EM over \pathfid in such cases where shortcuts may exist in the dataset and \pathfid relies on those shortcuts to get a higher answer EM but a lower support EM. Section \Scref{sec:dire_score} quantifies the extent to which \pathfid suffers from disconnected reasoning as compared to \model.   

\begin{figure*}[th!]
    \centering
    \includegraphics[width=0.99\textwidth]{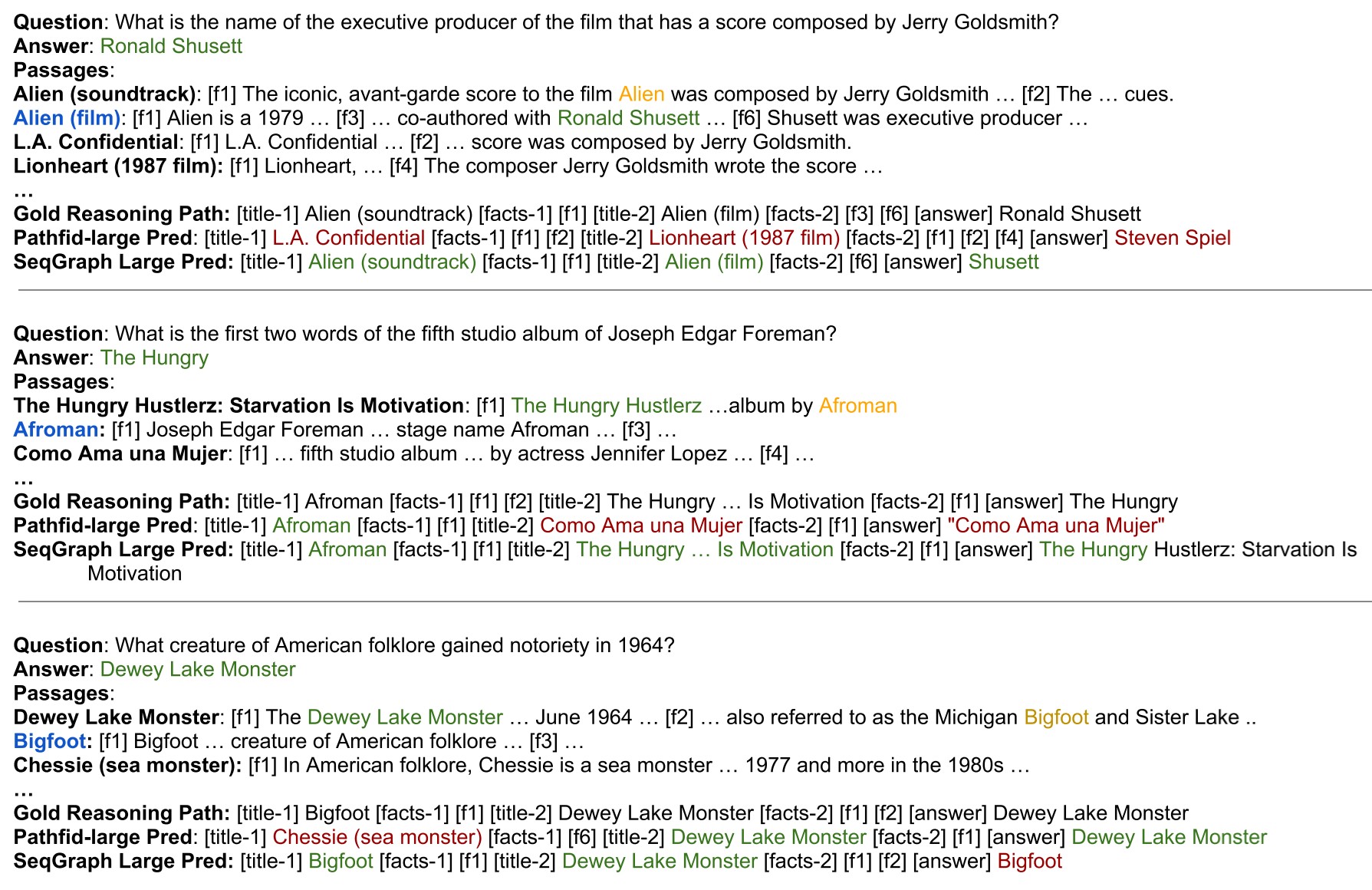}
    \caption{Qualitative Analysis of Disconnected Reasoning in \hotpot. \greentt{Correct}/\redtt{Incorrect} hops from \orangett{entity spans} to \bluett{Passage titles} for different cases are shown. In the first two cases, disconnected reasoning by \pathfid leads to incorrect final answer while \model gets the path and answer correct. The third case shows \pathfid getting the final answer right despite the reasoning path being disconnected while \model gets the connected reasoning path right.}
    \label{fig:dis_reasoning}
\end{figure*}

\subsection{Probing Disconnected Reasoning}
\label{sec:dire_score}

\hotpot suffers from information leakage in the form of reasoning shortcuts leading to \textit{disconnected reasoning}. This affects the generalization capability of such models and inflates the performance on the evaluation sets. Table \ref{fig:dis_reasoning} shows some qualitative examples of disconnected reasoning in \pathfid that are avoided by \model

\citet{trivedi-etal-2020-multihop} construct a probe of \hotpot by splitting the two supporting paragraphs for the original question across two questions. If the model can answer modified questions correctly without the complete context, it suggests that the model uses disconnected reasoning for the original question. By measuring the performance of a model on such a dataset, we arrive at the \textsc{DiRe} score with a higher value implying more disconnected reasoning. Table \ref{table:dire} shows the \textsc{DiRe} scores for the various models. We see that \model resorts to lower disconnected reasoning compared to \pathfid while maintaining strong performance gains on the original evaluation set.

\subsection{Comparison with PathFiD+}
\citet{yavuz-etal-2022-modeling} extend \pathfid and introduce \pathfid+ to improve the cross-passage interactions before feeding to the FiD decoder and show an improvement of ~7 EM points and achieve state-of-the-art results on \hotpot distractor dataset. However, we find the following limitations of the approach:

\textbf{Hop-assumption}: \pathfid+ adds pairs of contexts as input to the FID encoder, which assumes a fixed number of hops (in case of \hotpot, two) and doubles the input sequence length, leading to increased training time.

\textbf{Multi-step}: To efficiently encode pairs of passages (instead of inefficient \(\binom{N}{2}\) passages, where N is the total number of passages), \pathfid+ also needs to run the vanilla \pathfid or train another model to choose the first relevant context $P*$ to jump to and then construct pairs ($P*$, $P_n$). This makes it inefficient and not scalable to questions with higher hops or complex datasets like \musique

In contrast, our approach does not make any assumptions about the number of hops and is scalable. It produces output in a single shot without requiring multiple steps or increased sequence length. While \pathfid+ may achieve stronger performance in 2-hop \hotpot, our proposed method is more general, efficient and scalable, making it a more practical solution for real-world applications and also easily extendable to open-domain setting.

%% file: Tables/dire_score.tex
\begin{table*}[htbp!]
\centering
\small
\begin{adjustbox}{max width=\textwidth}
\begin{tabular}{rccccc}
\toprule \
\textbf{Model} &  \textbf{Answer} $\downarrow$  &  \textbf{Supp}$_\text{P}$ $\downarrow$  & \textbf{Supp}$_\text{S}$ $\downarrow$ & \textbf{Ans + }\textbf{Supp}$_\text{P}$ $\downarrow$  & \textbf{Ans + }\textbf{Supp}$_\text{S}$ $\downarrow$   \\
\midrule
\midrule 
\fid-Base & 51.1 & - & - & - & -   \\
\pathfid-Base & 45.5 & 48 & 49.1 & 22.6 & 24.3   \\
\model-Base & 44.7 & 46.2 & 45.4 & 21.8 & 22.8   \\
\midrule
\fid-Large & 53.5 & - & - & - & -   \\
\pathfid-Large & 48.8 & 48.3 & 49.7 & 24.3 & 26.4   \\
\model-Base & 45.7 & 45.9 & 45.3 & 22.3 & 23.4   \\
\bottomrule
\end{tabular}
\end{adjustbox}
\caption{\textsc{DiRe} score (F1 scores) for various models on the probe dataset of \hotpot indicating the extent of disconnected reasoning. Lower the score, the better the model.}
\label{table:dire}
\vspace{-4mm}
\end{table*}

%% file: sections/06-related.tex
\section{Related Works}

Multihop question answering requires a model to perform reasoning over multiple pieces of information, utilizing multiple sources and inferring relationships between them to provide a correct answer to a given question. There have been various approaches and datasets proposed for training QA systems, such as HotpotQA \cite{yang2018hotpotqa}, IIRC\cite{ferguson-etal-2020-iirc} and Musique \cite{10.1162/tacl_a_00475}.

In the \hotpot full-wiki setting, the task is to find relevant facts from all Wikipedia articles and then use them to complete the multi-hop QA task. Retrieval models play an important role in this setting, such as DPR~\cite{karpukhin-etal-2020-dense}, which focuses on retrieving relevant information in the semantic space. Other methods, such as Entities-centric~\cite{das-etal-2019-multi}, and Golden Retriever~\cite{https://doi.org/10.48550/arxiv.1910.07000}, use entities mentioned or reformulated in query keywords to retrieve the next hop document. Additionally, PathRetriever~\cite{asai2020learning} and HopRetriever~\cite{https://doi.org/10.48550/arxiv.2012.15534} use RNN to select documents to form a paragraph-level reasoning path iteratively. The above methods mainly focus on the open-domain setting (full-wiki) and improve the retriever's performance and do not address the disconnected reasoning problem.

Multiple techniques~\cite{jiang-bansal-2019-avoiding, lee-etal-2021-robustifying, ye-etal-2021-connecting} to counter disconnected reasoning operate
 at the dataset level, using adversarial training, adding extra annotations or using dataset augmentations to get a balanced train set and prevent the model from cheating.

We highlight differences between our approach and other related works on \hotpot-distractor and other works that combine language models with graphs below :

\textbf{Generative approaches}: Our generative-FiD approach differs from others using KG/GNN \cite{ju-etal-2022-grape, yu-etal-2022-kg} as we use an entity-passage graph with Wikipedia hyperlinks. Also, our focus is primarily on the distractor setting of multi-hop QA, while other baselines \cite{ju-etal-2022-grape, yu-etal-2022-kg} are either single-hop or improving retrieval in open-domain setting

\textbf{Pipeline vs single-stage}: Other baselines \cite{Tu2019SelectAA, Chen2019MultihopQA, qiu-etal-2019-dynamically, app11104699, 10096119} use a pipeline approach with distinct encoder models in the reasoning process, while we use a single-stage, one-shot prediction process without assumptions on the number of hops. 

\textbf{Graph construction}: Other methods \cite{Tu2019SelectAA, qiu-etal-2019-dynamically} select relevant passages heuristically from among distractors to construct graphs. However, we construct our entity-passage graph on all passages (including distractors) and fuse the representations in the encoder.

While a direct comparison with pipeline-based approaches is not possible or fair, we provide comparisons in Table \ref{tab:related-work-results} for completeness.

\input{Tables/related-work-comparison.tex}

%% file: Tables/related-work-comparison.tex
\begin{table}[htbp]
    \centering
    \begin{adjustbox}{width=\columnwidth}
        \begin{tabular}{@{}lcc@{}}
            \toprule
            Model & F1 & Support F1 \\
            \midrule
            DFGN\cite{qiu-etal-2019-dynamically} & 69.69 & 81.62 \\
            SAE-Large\cite{Tu2019SelectAA} & 80.75 & 87.38 \\
            \model-Base (T5-base) & 77.6 & 87.72 \\
            \model-Large (T5-large) & 81.62 & 88.28 \\
            C2FM-F1\cite{app11104699} (Electra large + DebertaV2 xx-large) & 84.65 & 90.08 \\
            FE2H\cite{10096119} (iterative Electra Large + Albert-xxlarge-v2) & 84.44 & 89.14 \\
            \bottomrule
        \end{tabular}
    \end{adjustbox}
    \caption{F1 scores of different related works on \hotpot distractor dataset}
    \label{tab:related-work-results}
\end{table}

%% file: sections/07-conclusion.tex
\section{Conclusion}
In this paper, we propose \model, an approach that utilizes the structured relationship between passages in the context of multi-hop questions to reduce disconnected reasoning. We construct a localized entity-passage graph using Wikipedia hyperlinks, encode it using a GNN, and fuse the structured representations with the text encoder for predicting a reasoning path. Our approach results in strong performance gains in terms of both answer and support EM/F1 on \hotpot and reduces disconnected reasoning measured using \textsc{DiRe} score. We also obtain state-of-the-art performance on the more challenging \musique benchmark with a 17-point improvement in answer F1 over the current best end-to-end(E2E) model. Experimenting with sophisticated methods of encoding the graph structure and fusing the text and graph representations can be explored in future work.

%% file: sections/appendix.tex
\begin{table}[h]
\small
\centering
\begin{adjustbox}{center}
\begin{tabular}{c c c c}
\toprule
Dataset & Train & Validation & Test \\ \midrule
HotpotQA - distractor & 90,447 & 7,405 & 7,405 \\ 
Musqiue - Answerable & 19,938 & 2,417 & 2,459 \\ \bottomrule
\end{tabular}
\end{adjustbox}
\caption{Number of samples in each data split for \hotpot and \musique.}
\label{tab:dataset-stats}
\end{table}

\section{Breakdown of Performance by Question Type - \hotpot}
\begin{table}[htbp]
\centering
\begin{tabular}{@{}r|c|c@{}}
\toprule
\textbf{Model} & \textbf{Bridge} & \textbf{Comparison} \\

\midrule
\fid-Base & 60.8 & 65.97  \\
\pathfid-Base & 61.19 & 65.37  \\
\model-Base & 63.6 & 66.51  \\
\hline
\pathfid-Large & 63.72 & 71.68 \\

\model-Large & 65.21 & 71.69 \\
\bottomrule
\end{tabular}
\caption{Performance breakdown of Answer-EM by question type on dev set of \hotpot}
\end{table}

\section{Reasoning Path variants in \musique}
\input{Tables/reason_path_variants.tex}

\input{Tables/musique-diff-pathfid-approaches.tex}

The different reasoning path variants that can be constructed based on ground truth annotations can be found in Table \ref{tab:Reasoning-path-variants}. Results of training baselines on these different variants can be found in Table \ref{tab:musique-diff-appraoches}

\section{Performance by Number of Hops - Graphs}

\begin{figure}[htbp]
    \centering
    \includegraphics[width=12cm, height=4cm]{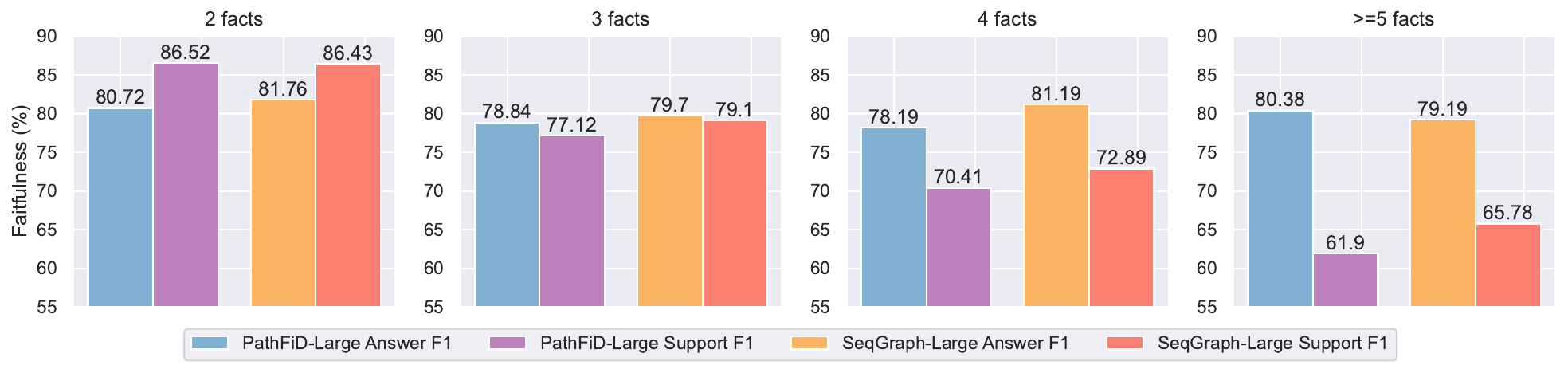}
    \caption{Performance on dev set of \hotpot decomposed by number of supporting facts.}
    \label{fig:hotpot_num_hops}
\end{figure}

\begin{figure}[H]
    \centering
    \includegraphics[width=12cm, height=3cm]{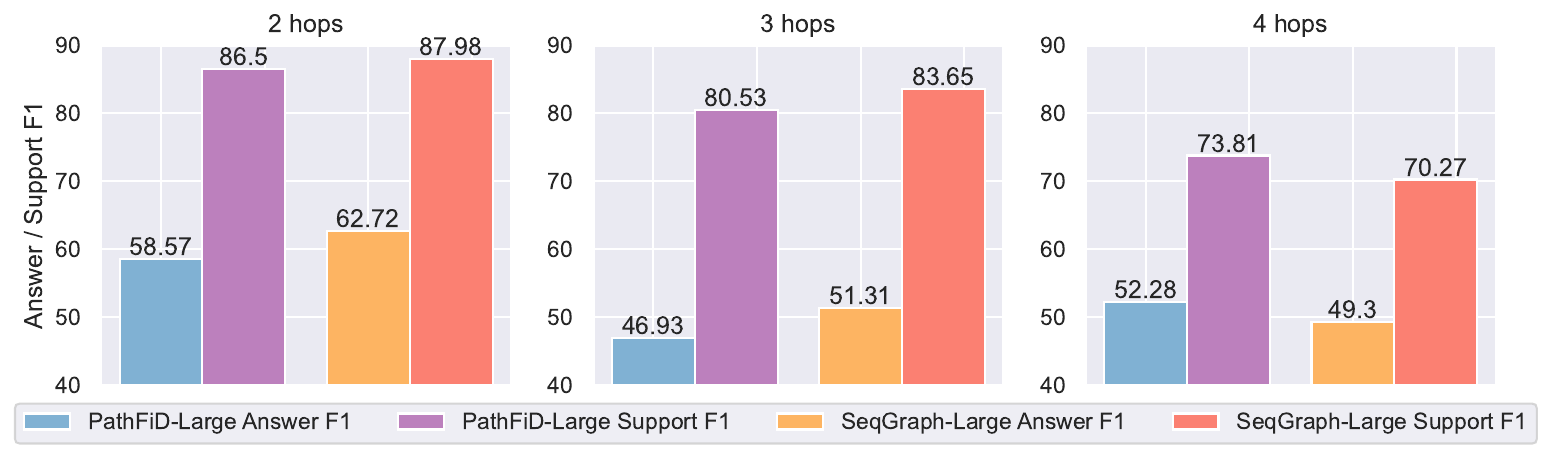}
    \caption{Performance on dev set of \musique decomposed by the number of hops.}
    \label{fig:musique_num_hops}
\end{figure}

We hypothesize that the answer F1 of \model on questions with $\geq 4$ hops gets impacted due to the presence of shortcuts since the support F1 score is higher than \pathfid.

\section{Comparison of Musique-Answerable test F1 scores}
Table \ref{tab:musique-leaderboard} shows the comparison of our models with the current best performing ones on the \musique- Answerable test set leaderboard. Our End-to-End single stage model \model-large trained to output title + intermediate answers (SIA) outperforms the Longformer-Large\cite{Beltagy2020Longformer} End-to-End model by 17 points in answer F1 and by 9-points in support F1. Furthermore, we also outperform the current state-of-the-art SA model which is a two stage model (Roberta Large\cite{Liu2019RoBERTaAR} + Longformer Large) by 5 points on Answer F1 and 3 points on Support F1.

\begin{table}[]
\small
\centering
\begin{adjustbox}{}
\begin{tabular}{l|cc}
\toprule
\textbf{Model} & \textbf{Answer F1} & \textbf{Support F1} \\ \hline
Select+Answer (SA) Model & 52.3 & 75.2 \\
Step Execution by Select+Answer (EX(SA)) Model & 49 & \textbf{80.6} \\
Step Execution by End2End (EX(EE)) Model & 46.4 & 78.1 \\
End2End (EE) Model & 40.7 & 69.4 \\
\hline
\fid-Large & 48.4 & XX \\
\pathfid-SIA-Large & 54.8 & 77.9 \\
\model-SIA-Large & \textbf{57.6} & 78.4 \\
\hline

\end{tabular}
\end{adjustbox}
\caption{Current best performing models on the leaderboard (Longformer-Large variants vs our baselines vs \model}
\label{tab:musique-leaderboard}
\end{table}

\section{Hyperparameter Settings}
\label{sec:detail-hyperparam}

Tables \ref{table:appendix_hparams_hotpot}, \ref{table:appendix_hparams_musique}, \ref{table: Additional Graph related params} detail the hyperparameters we use for \fid,\pathfid and \model for \hotpot and \musique.

The 2-layer GNN module is 17M parameters for the large model and 9.5M for the base, accounting for only upto 4\% increase in model parameters. 

\begin{table}[!t]
	\centering
	\scriptsize
	\begin{adjustbox}{max width=\textwidth}
		\begin{tabular}{lccc}
		    \hline  \\ [-2ex]
		    parameter & \textsc{\fid-Large} & \textsc{\pathfid-Large} \\
			\hline  \\ [-2ex]
			initialization &  t5-large &  t5-large \\
            learning rate &  1e-4  &  1e-4 \\
            learning rate schedule & linear & linear  \\
            effective batch size &  64 &  64 \\
            gradient checkpointing &  yes &  yes \\
            maximum input length   & 256 & 256 \\
            maximum output length  & 32 & 64 \\
            warmup steps           & 2000 & 2000 \\
            gradient clipping norm & 1.0 & 1.0 \\
            training steps         & 16000 & 16000 \\
            weight decay           & 0.01 & 0.01 \\
            optimizer              & adamw & adamw \\
			\hline
		\end{tabular}
	\end{adjustbox}
	\vspace{1mm}
	\caption[Table caption text]{Hyperparameters for experiments on HotpotQA Distractor setting.}
	\label{table:appendix_hparams_hotpot}
\end{table}

\begin{table}[!t]
	\centering
	\scriptsize
	\begin{adjustbox}{max width=\textwidth}
		\begin{tabular}{lccc}
		    \hline  \\ [-2ex]
		    parameter & \textsc{\fid-Large} & \textsc{\pathfid-Large-SIA} \\
			\hline  \\ [-2ex]
			initialization &  t5-large &  t5-large \\
            learning rate &  1e-4  &  1e-4 \\
            learning rate schedule & linear & linear  \\
            effective batch size &  64 &  64 \\
            gradient checkpointing &  yes &  yes \\
            maximum input length   & 256 & 256 \\
            maximum output length  & 32 & 90 \\
            warmup steps           & 1000 & 1000 \\
            gradient clipping norm & 1.0 & 1.0 \\
            training steps         & 6500 & 6500 \\
            weight decay           & 0.01 & 0.01 \\
            optimizer              & adamw & adamw \\
			\hline
		\end{tabular}
	\end{adjustbox}
	\vspace{1mm}
	\caption[Table caption text]{Hyperparameters for experiments on Musique-Answerable setting.}
	\label{table:appendix_hparams_musique}
\end{table}

\begin{table}[!t]
	\centering
	\scriptsize
	\begin{adjustbox}{max width=\textwidth}
		\begin{tabular}{lccc}
		    \hline  \\ [-2ex]
		    parameter & \textsc{\model-Large} \\
			\hline  \\ [-2ex]
			GNN                  & GAT\cite{https://doi.org/10.48550/arxiv.1710.10903} \\
                GNN Hidden Dimension &  1024 \\
                GNN Number of layers & 2 \\
                GNN dropout          & 0.2 \\
                Number of heads      & 8 \\
                Layer for fusion $L$ & 3 \\
			\hline
		\end{tabular}
	\end{adjustbox}
	\vspace{1mm}
	\caption[Table caption text]{Additional Graph related hyperparameters for SeqGraph}
	\label{table: Additional Graph related params}
\end{table}

%% file: Tables/reason_path_variants.tex
\begin{table}[htbp]
\tiny
\resizebox{\textwidth}{!}{%
\begin{tabular}{ll}
\toprule[1.5pt]
HotpotQA &                \\ \cmidrule{1-1}
Question:      & What is the name of the executive producer of the film that has a score composed by Jerry Goldsmith?                                  \\
Answer:  & Ronald Shusett \\
Reasoning Path: & {[}title-1{]} \bluett{Alien (soundtrack)} {[}facts-1{]} \orangett{[f1]} {[}title-2{]} \bluett{Alien (film)} {[}facts-2{]} \orangett{[f6]} {[}answer{]} \greentt{Ronald Shusett} \\ \midrule[1.5pt]
Musique & \\ \cmidrule{1-1}
Question: &              Who is the spouse of the Green performer? \\
Answer: &                 Miquette Giraudy \\
Reasoning Path: & \\ 
DA: &   [question-1] \browntt{Who is the performer of Green?} [question-2] \browntt{Who is the Spouse of \#1?} [answer] \greentt{Miquette Giraudy} \\
SA: &    [title-1] \bluett{Green (Steve Hillage album)} [title-2] \bluett{Miquette Giraud} [answer] \greentt{Miquette Giraudy}\\
SIA:&    [title-1] \bluett{Green (Steve Hillage album)}  [answer-1] \greentt{Steve Hillage} [title-2] \bluett{Miquette Giraudy} 
 [answer] \greentt{Miquette Giraudy} \\

 \multirow{2}{*}{DSIA:} & [question-1] \browntt{Who is the performer of Green?}   [title-1] \bluett{Green (Steve Hillage album)}  [answer-1] Steve Hillage    \\ 
 &  [question-2] \browntt{Who is the Spouse of \#1?}  [title-2] \bluett{Miquette Giraudy} [answer] \greentt{Miquette Giraudy} \\\bottomrule
\end{tabular}%
}
\caption{Reasoning path variants for \hotpot and \musique. Relevant passage titles are marked in \bluett{blue}, supporting facts in \orangett{orange}, intermediate answer/final answer is marked in \greentt{green} and the decomposed questions are marked in \browntt{brown}}
\label{tab:Reasoning-path-variants}
\end{table}

%% file: Tables/musique-diff-pathfid-approaches.tex
\begin{table}[htbp]
    \centering
    \begin{tabular}{@{}lcccc@{}}
        \toprule
        Model & Answer-EM & Answer-F1 & Support-EM & Support-F1 \\
        \midrule
        SA & 32.02 & 41.76 & 47.04 & 76.23 \\
        DA* & 31.61 & 41.4 & XX & XX \\
        SIA & 34.71 & 44.93 & 57.3 & 80.18 \\
        DSIA & 33.35 & 43.08 & 53.5 & 78.79 \\
        \bottomrule
    \end{tabular}
    \caption{Results on different variants of \musique reasoning paths. *Since DA does not predict a reasoning path with titles, we do not compute the Support EM and F1.
}
    \label{tab:musique-diff-appraoches}
\end{table}